\documentclass[10pt,conference]{IEEEtran}
\IEEEoverridecommandlockouts
\usepackage{cite}
\usepackage{amsmath,amssymb,amsfonts}
\usepackage{booktabs}
\usepackage{algorithmic}
\usepackage{graphicx}
\usepackage{textcomp}
\usepackage{xcolor}
\def\BibTeX{{\rm B\kern-.05em{\sc i\kern-.025em b}\kern-.08em
    T\kern-.1667em\lower.7ex\hbox{E}\kern-.125emX}}
\begin{document}

\title{Design and Prototyping Distributed CNN Inference Acceleration in Edge Computing}

\author{\IEEEauthorblockN{Zhongtian Dong, Nan Li, Alexandros~Iosifidis, and Qi Zhang }
\IEEEauthorblockA{\textit{DIGIT, Department of Electrical and Computer Engineering, Aarhus University} \\
Email: dzt97321@gmail.com, \{linan, ai, qz\}@ece.au.dk}}

\maketitle

\begin{abstract}
For time-critical IoT applications using deep learning, inference acceleration through distributed computing is a promising approach to meet a stringent deadline. In this paper, we implement a working prototype of a new distributed inference acceleration method HALP using three raspberry Pi 4. HALP accelerates inference by designing a seamless collaboration among edge devices (EDs) in Edge Computing. We maximize the parallelization between communication and computation among the collaborative EDs by optimizing the task partitioning ratio based on the segment-based partitioning. Experimental results show that the distributed inference HALP achieves $1.7 \times$ inference acceleration for VGG-16. 
Then, we combine distributed inference with conventional neural network model compression by setting up different shrinking hyperparameters for MobileNet-V1. In this way, we can further accelerate inference but at the cost of inference accuracy loss. To strike a balance between latency and accuracy, we propose dynamic model selection to select a model which provides the highest accuracy within the latency constraint. It is shown that the model selection with distributed inference HALP can significantly improve service reliability compared to the conventional stand-alone computation.
\end{abstract}

\section{Introduction}
Today, artificial intelligence based on deep convolutional neural networks (CNNs) is an integral part
of our everyday life, ranging from autonomous driving, AR/VR, and smart surveillance to virtual assistants such as Alexa, Siri, and Google Assistant. However, IoT and mobile devices have limited processing capabilities and battery lifetime, whereas CNN inference is computationally expensive. This makes IoT devices often unable to run CNN inference or meet the latency requirements of time-critical IoT applications \cite{Nan2022}. To address this issue, the conventional approach is to offload the inference task to the Cloud. However, due to long communication latency, Cloud Computing cannot meet latency constraints. To this end, Edge Computing is a promising computing paradigm, which allows an IoT device to offload computation tasks to nearby computing resources \cite{LiuIoT:2020}. Although the computation resources at network edge or in proximity are not as powerful as those at the Cloud, due to short communication latency, many recent works have shown the benefits of Edge Computing for time-critical IoT applications~\cite{Jianhui2020Access}. 

There exist different techniques to accelerate inference, such as model compression, model partitioning, task partitioning using distributed inference, dynamic inference e.g. early exit, and others. Model compression is to compress a CNN model into a lightweight model through parameter pruning or quantization, which could lead to slight inference accuracy loss. Model partition, also called split computing or partial offloading, is to offload the computationally intensive parts of a CNN model to an edge server. The key issue of model partition is that the intermediate output feature maps (i.e. tensor) of a CNN could be larger than the input image. Early exiting allows early inference termination to meet the deadline by slightly sacrificing inference accuracy~\cite{ArianEarlyexit:2021, bakhtiarnia2022dynamic}. Task partitioning using distributed inference is to partition task data (e.g. an image) into several sub-tasks and offload each sub-task (e.g. a segment of an image) to a collaborative device (or server) and all devices (or servers) collaboratively complete the inference. These techniques are complementary, therefore, it is possible to combine them to further accelerate inference, as will be shown in Section~\ref{Sec:PerEva}.
%Distributed DNN is a new technique proposed in recent years, which accelerates inference time through reasonable partition and offloading of inference tasks.

Segment-based partitioning~\cite{Mohammed2020:Distributed} partitions the input tensor of each convolutional layer (CL) into small segments along the largest dimension (e.g. along width or height) to accelerate inference. However, it leads to accuracy loss caused by the effect of stride and padding on each convolutional layer. To address this issue, a new distributed inference method called Host Assisted Layer-wise Parallelization (HALP)~\cite{Nan2022ICC} proposed using receptive field-based segmentation \cite{Nan2022}. Furthermore, HALP parallelizes the computation and communication processes among collaborative Edge servers through a seamless collaboration scheme. Simulation results show that HALP can achieve $1.7-2.0 \times$ speedup for VGG-16 on GTX 1080TI and JETSON AGX Xavier with 9 collaborative edge servers at communication data rate of 100 Gbps~\cite{Nan2022ICC}. However, it has not yet been evaluated in real edge computing scenarios, where computation time and data rate could be time-varying. This paper aims to implement a working prototype of the distributed inference HALP and validate inference acceleration performance. Furthermore, this paper combines the prototype of HALP with model compression methods to study the trade-off between latency and accuracy. In addition, this paper studies how to strike a balance between latency and accuracy using dynamic model selection under a latency constraint. 
\begin{figure*}[t]
\centering
\begin{minipage}[]{0.38\textwidth}
\centering
\includegraphics[width=1\textwidth]{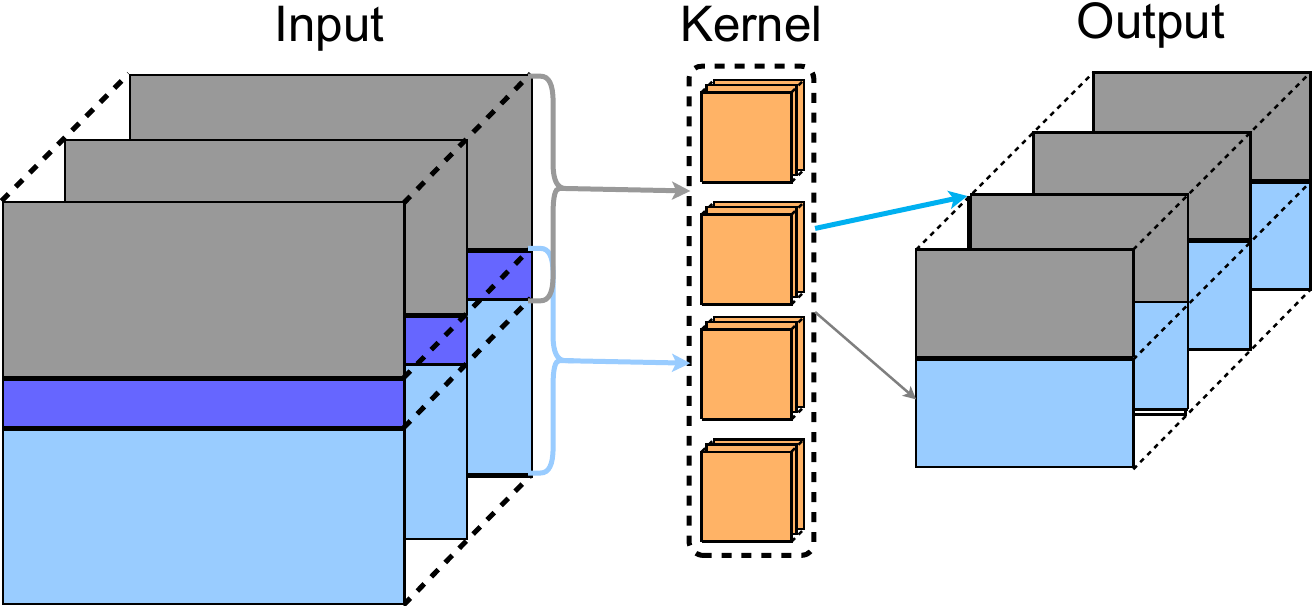}
\caption{Segment-based partitioning for distributed inference}\label{Fig:segment}
\end{minipage}
\hspace{1mm}
\begin{minipage}[]{0.58\textwidth}
\centering
\includegraphics[width=1.0\textwidth]{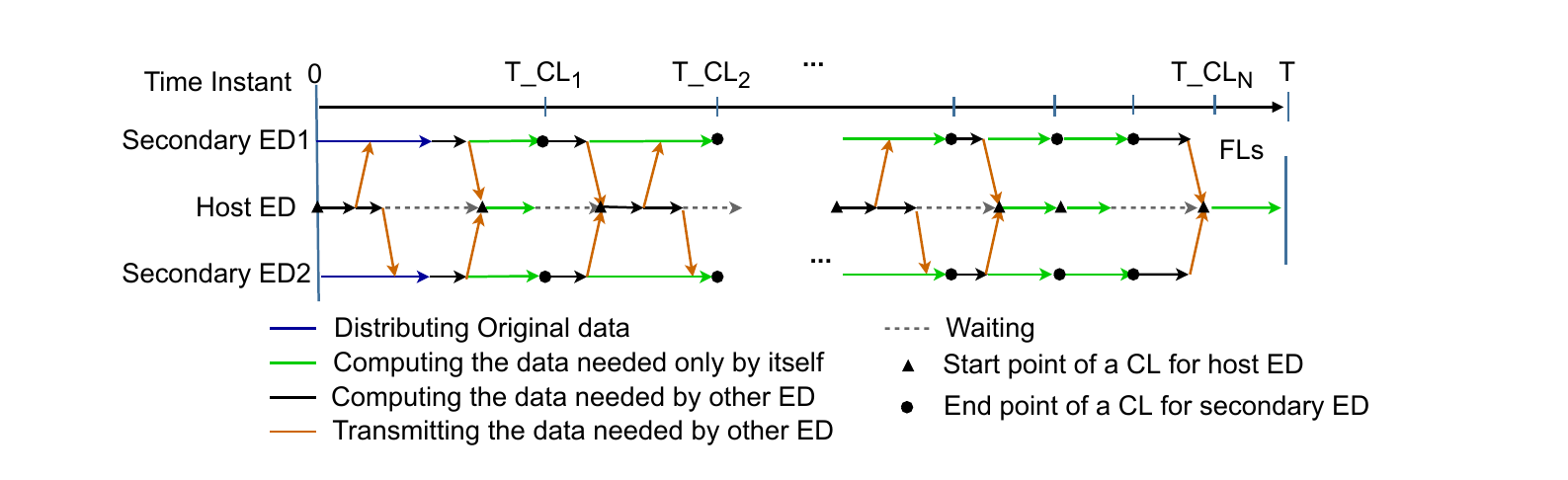}
\vspace{-5mm}
\caption{Illustration of communication and computation in HALP for VGG-16}\label{Fig:halp}
\end{minipage}
\end{figure*}

Our main contributions are summarized in the following. (i) We implement a working prototype of distributed inference HALP for both VGG-16 and MobileNet-V1 using three Raspberry Pi 4. (ii) We optimize task partitioning in HALP, achieving $1.7 \times$ inference acceleration for VGG-16. (iii) We combine distributed inference HALP with CNN model compression using various shrinking hyperparameters for MobileNet-V1 to further improve inference acceleration but at the cost of inference accuracy loss. This illustrates the complementary characteristics of distributed inference HALP and model compression methods. It also shows a clear trade-off between latency and accuracy. (iv) We propose dynamic model selection to select a model which can achieve the highest accuracy within a given latency constraint, taking into account the inference time of different models and the transmission time based on the current data throughput. Experimental results show that dynamic model selection with distributed inference HALP achieves higher service reliability in comparison with stand-alone computation. 

This paper is organized as follows: Section~\ref{Sec:Background} briefly describes how HALP works. Section~\ref{Sec:SystemModel} shows the system architecture of our prototype. Section~\ref{Sec:PerEva} presents our experimental results on inference time using the implemented prototype of distributed inference, and it shows how dynamic model selection with distributed inference can improve service reliability. %Section V illustrates our dynamic model selection method and corresponding evaluation for the Quality of service with and without HALP. 
We conclude this paper in Section~\ref{Sec:Conclusion}.

\section{Background}\label{Sec:Background}
Conventional segment-based partitioning method partitions an inference task into sub-tasks, and each sub-task is offloaded from a host to a secondary edge device (ED), as shown in Fig.~\ref{Fig:segment}. However, due to the effect of padding and stride, information at the boundary between sub-tasks is lost. This may lead to accuracy loss. For example, if the outputs of sub-tasks at each CL are not exchanged, for the VGG-16 which has a kernel size of $3 \times 3$, and padding and stride equal to 1, the output of each sub-task will not consider one line at its boundary from its adjacent sub-task. To address this issue, HALP uses receptive-field in segment-based partitioning to calculate the exact output for a CNN model without compromising inference accuracy. In addition, HALP leverages seamless collaboration in edge computing to maximize the parallelization between communication and computing processes, thereby minimizing the total inference time of an inference task. 

The basic idea of HALP is as follows. (i) Before starting the computation of the first CL, the host first partitions an image into three parts. Two of them are close to half of the image size, and each of them is sent to a collaborating secondary ED. The third part (i.e. the overlapping zone) is processed by the host itself. Note that the overlapping zone is often very small in CNNs. For example, for VGG-16, the overlapping zone is only 4 rows of pixels, since the kernel size used in VGG-16 is $3 \times 3$. (ii) The host performs the computation of the overlapping zone of the first CL, while it is transmitting parts of the original input image to the secondary EDs. Then, the host sends the corresponding output of the first CL's overlapping zone to the secondary EDs once it is available. Each secondary ED, after receiving the part of the original image, first computes the part of the first CL's output needed by the host to proceed with the computation of the second CL. A secondary ED computes the rest output of the first CL needed only by itself and transmits the part needed by the host to the host. Such seamless collaboration between the host and secondary EDs repeats until the last CL is completed. Then all sub-outputs of the secondary EDs are sent to the host and then are merged as the input for the fully-connected layers (FLs). The communication and computation process in HALP is also illustrated in Fig.~\ref{Fig:halp}.
\begin{figure}[t]
  \centering
  % include first image
  \includegraphics[width=0.65\linewidth]{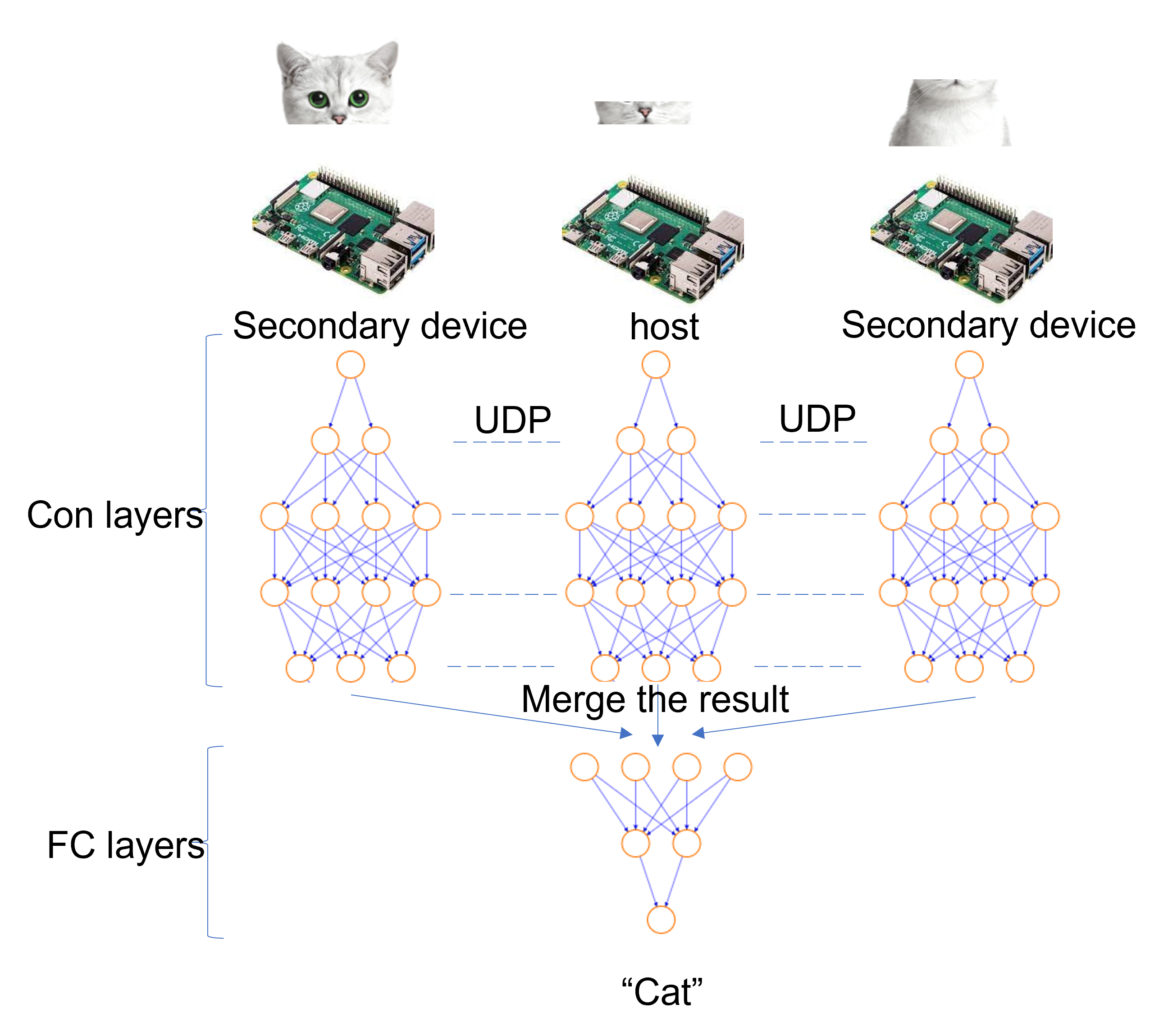}  
 \caption{System overview of distributed inference HALP prototype}\label{Fig:SystemOverview}
 % \label{fig:ieee_float_format_vector}
\end{figure}

\section{System Model}~\label{Sec:SystemModel}
We consider an edge computing network with three collaborative EDs, one of which is the host ED and the other two are secondary EDs.
We implement a working prototype of distributed inference HALP, as shown in Fig.~\ref{Fig:SystemOverview}.
\subsection{Task partitioning}
We implement task partitioning for two CNN models, VGG-16~\cite{Karenvgg:2015} and MobileNet-V1~\cite{AndrewMobileNet:2017} with the input image size of $224 \times 224 \times 3$ (i.e. height, width, channel). The host ED first partitions the input image along the height dimension into two equal parts then offloads the top half to the first secondary ED (ED1) and the bottom half to the second secondary ED (ED2). The host ED processes the overlapping zone (4 rows) needed by the secondary EDs to proceed with the next CL.

%An image is first partitioned along the height into two equal parts, and the upper part is offloaded to the first secondary ED (ED1), the lower part is offloaded to the second secondary ED (ED2), the host ED (ED0) takes the 4 rows in the middle which is the overlapping part, meaning that each secondary ED needs the output of the overlapping part for the next CL.

\subsubsection{VGG-16 task partitioning}
VGG-16 has 5 blocks. % with a pooling layer at the end of each block. For each block, the height $H$ and width $W$ of the output tensor will become half of the input, the number of channel $C$ remains the same. 
Each block contains several convolution layers and a pooling layer, and in the same block, the number of channels in the convolution layers is same. Convolution process is basically to perform a dot product between a kernel or filter and the patches of input data (same as kernel size). Stride defines how many pixels a kernel moves for each step. Pooling layer performs by reducing the tensor size. %Convolution layer preserves the relationship between pixels by learning image features using small patches of input data. Stride defines how many pixels should the kernel moves, pooling layer performs downsampling by reducing the size and sends only the important data to next layers in CNN.
Note that the secondary EDs do not need any information from the host to proceed with the pooling layer because the secondary EDs already have the information. The detailed input rows of each block are shown in Table~\ref{tab:partition_vgg16}.
%VGG-16 has 5 blocks, and each block ends up with a pooling layer after maximum pooling, the height $H$ and width $W$ of the output tensor will become half of the original, the number of channel $C$ remains the same. One thing to be noticed is that, before the pooling layer, the secondary EDs do not need any information from the host because before pooling layer, the secondary EDs only need $(H-2)\times W\times C$ input, after maximum pooling operation, the $H$ then becomes $\frac{H-2}{2}$, plus one row from the host ED($\frac{H}{2}$), the secondary EDs already have the information to process the next convolution layer. TABLE II shows
\begin{table}[]
	\centering
	\begin{minipage}[b]{1.0\linewidth}
	\caption{Task partitioning for VGG-16}~\label{Tab:partitionVGG}
	\centering
	\resizebox{0.85\textwidth}{!}{
		\begin{tabular}{c|ccc|ccc}
			\toprule
		Block	& \multicolumn{3}{c|}{Default partition} & \multicolumn{3}{c}{Optimized partition} \\
			& \multicolumn{1}{c}{Host} & \multicolumn{1}{c}{ED1} & \multicolumn{1}{c|}{ED2} 	& \multicolumn{1}{c}{Host} & \multicolumn{1}{c}{ED1} & \multicolumn{1}{c}{ED2} \\
			\cline{1-7}
  Block1        & 4 & 112 &   112 & 68 & 80 &   80\\ 
  Block2        & 4 & 56 &   56  & 36 & 40 &   40 \\
  Block3        & 4 & 28 &   28  & 20 & 20 &   20\\
  Block4        & 4 & 14 & 14 & 12 & 10 & 10 \\
  Block5        & 4 & 7 & 7 & 8 & 5 & 5\\
			\bottomrule
	\end{tabular}}%
	\label{tab:partition_vgg16}
\end{minipage}
\end{table}%

For the default partitioning, the host ED only takes the overlapped zone (4 rows) of an input task and the secondary EDs need more time for computing the sub-tasks than the host. Therefore, the host ED has to wait for the secondary EDs to complete the computation before processing the next CL. In other words, the parallelization between EDs can be further improved by increasing the size of the overlapped zone. Taking into account the communication data rate and computing time of each sub-task and the overlapping zone, we can optimize task partitioning to minimize the idle time of the host ED.

Since the stride of VGG-16 is always 1, for each block the height and width of the output feature maps of CLs remain unchanged, and the output of the pooling layer is halved. For the host, the number of rows of the overlapped zone is half of the number of rows in the previous layer plus two rows from the secondary EDs. Hence, the number of rows in the overlapped zone of block $i$, $z_i$, is given as:
\begin{equation}
z_i=\frac{z_{i-1}}{2}+2     .
\end{equation}

%There are total 5 blocks in VGG-16 and the last layer of each block is a pooling layer. Since the stride of VGG-16 is always 1, the height and width of an image will not change in all layers except the pooling layer. After the max pooling operation, the height and width of the image are adjusted to half of the original, and the number of channels of the picture remains the same. For the host ED, the number of rows of the overlapped part is half of the number of rows in the previous layer plus two rows from the secondary EDs. Formally, the number of rows in the overlapped part of block $i$, $r_i$, is given as:

If $z_1=4$, the number of rows in all the blocks is always equal to 4. There are several principles to follow when selecting the number of rows. (i) This number should be divisible by 2 so that the output of max-pooling is reasonable. (ii) The number of rows in the secondary ED is also dependent on the number of rows in the host ED. The number of rows distributed on the three EDs should be approximately equal at each layer in order to minimize the idle time of the host ED.

We start to look at the last block, which has an input feature map with a height of 14. Except for 4, we can choose from $\{6,~8,~10,~12\}$. For $10$ and $12$, the proportion of tasks assigned to the host is too large. For 6, the number of rows of the secondary EDs is an odd number which violates the rule of pooling. Therefore, 8 is the best choice. The optimized task partitioning is shown in Table~\ref{Tab:partitionVGG}.
\subsubsection{MobileNet-V1 task partitioning}
MobileNet-V1 replaces the standard convolutional layer in VGG-16 with depth-wise separable convolution which consists of depth-wise convolution and point-wise convolution. The partitioning of the MobileNet-V1 task is slightly different from VGG-16 due to varying stride size. Except for the average pooling layer before the fully-connected layer, MobileNet-V1 does not have any other pooling layer and it performs down-sampling of feature maps by setting the stride of some CLs to 2. When the stride is 1 and the kernel size of a CL is $3 \times 3$, the host ED needs to compute $4$ rows for the data needed for the secondary EDs. If the stride is 2, then the kernel or filter is moved $2$ pixels at a time. So for this case, the host ED needs to compute $5$ rows.

In summary, for depth-wise convolution, when the stride is $2$, the host ED computes 5 rows, and when the stride is $1$, the host ED computes $4$ rows which are the same as in VGG-16. Note that before each layer with stride $2$, the first secondary ED needs to send the last row to the host ED so that it will have 5 rows to compute. For point-wise convolution, since the kernel size is $1 \times 1$, there is no need for data exchange except in the situation mentioned above. We describe the detailed partition of an inference task for MobileNet-V1 in Table \ref{tab1}.
\begin{table}[]
	\centering
	\begin{minipage}[b]{1.0\linewidth}
	\caption{Task partitioning for MobileNet-V1}
	\centering
	\resizebox{0.65\textwidth}{!}{
		\begin{tabular}{c|ccc}
			\toprule
		Layers & \multicolumn{1}{c}{Host} & \multicolumn{1}{c}{ED1} & \multicolumn{1}{c}{ED2} 	\\
			\cline{1-4}
  Conv /s2        & 5 & 112 &   112\\ 
  Conv dw/s1        & 4 & 112 &   112  \\
  Conv dw/s2        & 5 & 56 &   56  \\
  Conv dw/s1       & 4 & 56 & 56  \\
  Conv dw/s2       & 5 & 28 & 28  \\
  Conv dw/s1       & 4 & 28 & 28  \\
  Conv dw/s2       & 5 & 14 & 14  \\
  Conv dw/s1 x5       & 4 & 14 & 14  \\
  Conv dw/s2       & 5 & 7 & 7  \\
  Conv dw/s1        & 4 & 7 & 7  \\
\bottomrule
\end{tabular}}%
\label{tab1}
\end{minipage}
\end{table}%

\subsection{Inference and transmission scheduling}
\subsubsection{VGG-16 task scheduling}
The host ED first transmits parts of an image to the secondary EDs. Note that to reduce transmission time, based on the data size or format of the image, the host can choose to transmit either directly the image or the representation of the image, such as a tensor (the unit of a tensor is a float32 number). For VGG-16, the input is a tensor of $224 \times 224 \times 3$. When offloading half of the input tensor to secondary EDs, the transmission data size can be calculated as $112 \times 224 \times 3 \times 32/\left(1024\times 8\right)= 294 \text{Kbits}$. If the input image size is smaller than $294~\text{Kbits}$, it is better to directly offload the whole image to the secondary EDs; otherwise, it is better to offload half of the input tensor to the secondary EDs. To maximize parallelization between communication and computation, the host ED and secondary EDs perform the inference by following the process of HALP, which has been described in Section \ref{Sec:Background}.

\subsubsection{MobileNet-V1 task scheduling}
The scheduling of inference task in MobileNet-V1 is quite similar to that of VGG-16. Most of the time, only intermediate data in the depth-wise CL is exchanged. Because the kernel size of the point-wise CL is $1 \times 1$, after the data exchange of the previous depth-wise CL is completed, the three EDs are able to compute the next point-wise CL without any data exchange. Note that when the stride of the depth-wise CL is $2$, the first secondary ED needs to send one row data to the host ED after the point-wise CL. This is because the output of the previous point-wise CL at the host ED is only 4 rows and the host ED needs 5 rows to proceed with the next depth-wise CL computation. 
%means that the amount of data exchange will increase compared to the depth-wise convolution layer. The increase in the amount of data means increasing transmission time. This increase may affect the total inference time, when the channel state is not good.
%There is a special case where the first secondary ED needs to send data to the host ED in the point-wise CL, that is, when the stride of the next layer is equal to $2$, because the host needs $5$ lines to compute the data required by the secondary ED2 and the output of each layer of the host is $4$ rows, so the 1st secondary ED needs to send one row to the host in the point-wise CL.

\section{Performance Evaluation}~\label{Sec:PerEva}
In this section, we first describe the experimental setup, then we show the impact of task partitioning on the inference acceleration using VGG-16 as an example. Furthermore, we combine distributed inference with CNN model compression for MobileNet-V1 and show how model compression can further accelerate inference. Moreover, we evaluate how the dynamic model section with distributed inference can maximize the service reliability by striking a good balance between latency and accuracy under the latency constraints.

\subsection{Experiment setup}
We develop a working prototype of distributed inference HALP which works over WiFi. The inference acceleration depends on the communication data rate. The higher the computation capacity of the collaborative EDs, the greater the communication data rate is needed to achieve inference acceleration. We use Raspberry Pi 4 as our EDs, which communicate with each other using UDP (User Datagram Protocol) which has low latency. We use the ImageNet dataset. The measured data throughput $r$ between EDs varies between 26 Mpbs to 52 Mpbs with an average of 42 Mbps. Note that the measured data throughput here includes the data transmission time, as well as the time to serialize a tensor at the sender and the time to deserialize the tensor at the receiver.
\begin{figure}[t]
  \centering
  % include first image
  \includegraphics[width=0.8\linewidth]{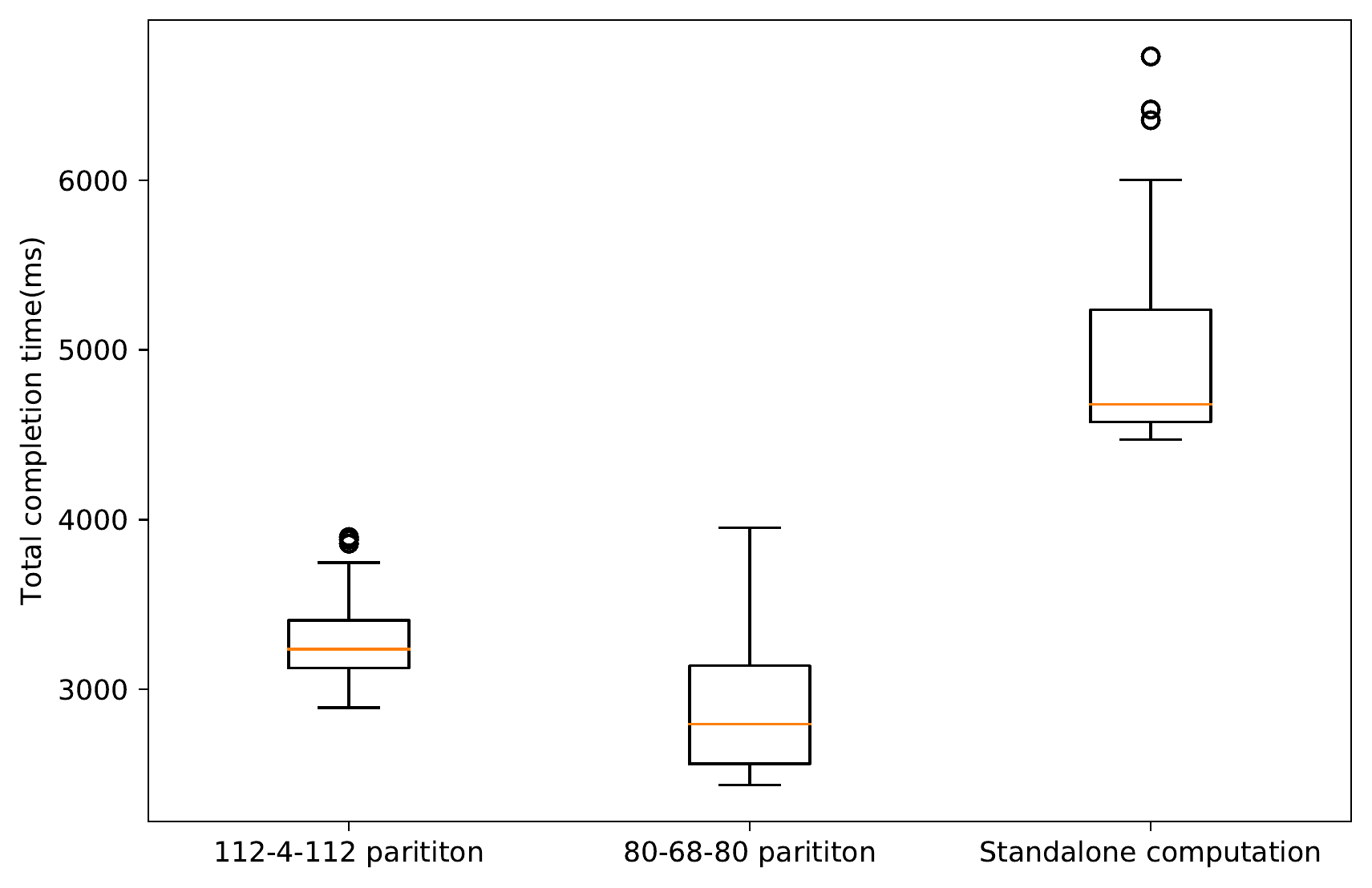}  
 \caption{Inference time of distributed inference for VGG-16 using three Raspberry Pi 4, $r=42~\text{Mbps}$}\label{Fig:InferenceTimeVGG}
  \label{fig:ieee_float_format_vector}
\end{figure}

\subsection{Task partitioning in VGG-16}
Fig.~\ref{Fig:InferenceTimeVGG} shows the overall inference time to perform  inference by VGG-16 using different partition methods, For each partition method, the experiment is repeated 10,000 times. The partition `112-4-112' means that each secondary ED takes a sub-task of $112$ rows and the host takes the overlapping zone of $4$ rows. It can be seen that the inference time is not static. For example, on a stand-alone Raspberry Pi 4, the inference time varies between 4472 ms and 6730 ms with an average of 4905 ms. This results from fluctuations in computation time and communication time caused by time-varying channel states. The inference time of distributed inference is clearly shorter compared with that of a stand-alone computation. For example, the inference time of partition `112-4-112' varies between 2891 ms and 3857 ms with an average of 3264 ms. By optimizing the task partitioning, the inference time can be further reduced to between 2435 ms and 3953 ms with an average of 2864 ms with the optimized partition `80-68-80'. Moreover, we can see that the worst case of distributed inference HALP still outperforms the best case of stand-alone computation. 

We define the acceleration gain as the inference time of stand-alone computation divided by that of HALP. Table~\ref{Tab:AccelerationGain} summarizes the average inference time and the gain of two different partitioning methods. We can see the optimized partitioning method achieves an acceleration up to $1.7 \times$. %which indicate that balancing computing resource between devices do improve the computing parallelism.
\begin{table}[]
	\centering
	\begin{minipage}[b]{1.0\linewidth}
	\caption{Inference time and acceleration gain of distributed inference in Raspberry Pi 4 for VGG-16, $r=42$ M{\upshape bps}}
	\centering
	\resizebox{0.9\textwidth}{!}{
		\begin{tabular}{c|ccc}
			\toprule
		Method & Average inference time (ms) & Gain\\ 
			\cline{1-3}
Standalone computation       & 4905 & 1 \\ 
Partition 112-4-112       & 3264 & 1.50 \\ 
Partition 80-68-80       & 2864 & 1.71 \\ 
\bottomrule
\end{tabular}}%
\label{Tab:AccelerationGain}
\end{minipage}
\end{table}%

\subsection{Task partitioning with CNN model compression for MobileNet-V1 } \label{model compression}
To accelerate CNN inference, a lightweight model can often be used to reduce the multiply–accumulate operations in convolution. To realize this, \cite{AndrewMobileNet:2017} proposed two methods: 1) using width multiplier $\alpha$ to reduce the number of channels by $1-\alpha$ at each layer; 2) using resolution multiplier $\rho$ to resize the resolution (height and width) of an input image.

\begin{figure}[t]
  \centering
  % include first image
  \includegraphics[width=0.9\linewidth]{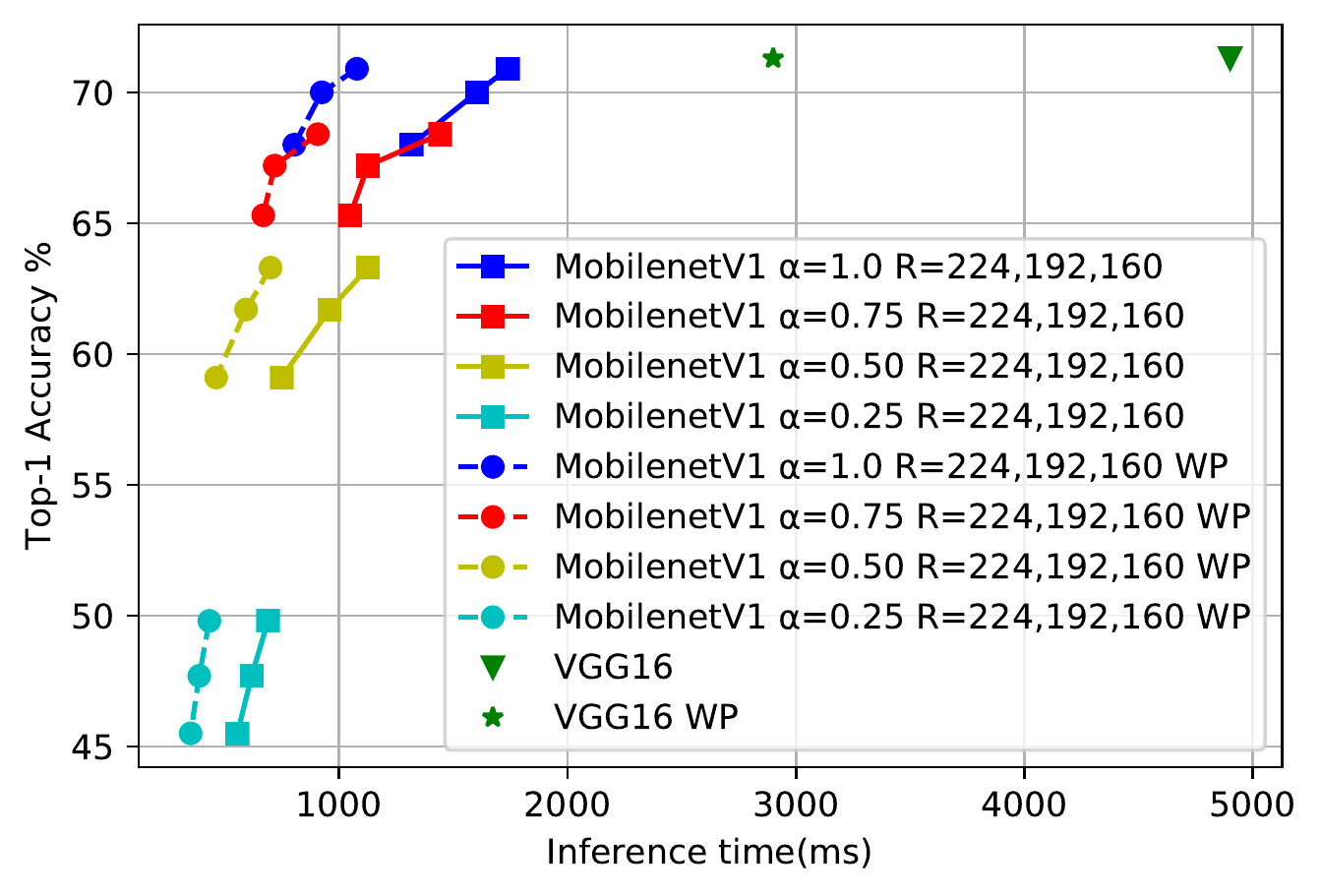} 
 \caption{The trade-off between inference time and accuracy on raspberry Pi 4, $r=42$ Mbps (`WP': with partition)}\label{Fig:accuracyVStime}
  \label{fig:ieee_float_format_vector}
\end{figure}

We conduct the tests of 12 lightweight MobileNet-V1 models generated from the combination of the width multiplier $\alpha \in \{1, 0.75, 0.5, 0.25\}$ and the resolutions $\{224\times 224, 192\times 192, 160\times 160\}$. Table~\ref{Tab:GainMobileNet} shows performance of the 12 lightweight models using HALP. We can see the lightweight models can significantly accelerate the inference.
%The inference time of MobileNet-V1 is much shorter than that of VGG16, and the gain is generally higher than that of VGG16. This is because the data is aggregated in the fully connected layer, and then computed only by the host \acrshort{ED, and the computation of the fully connected layer of VGG16 is about 8\% of the total inference time, and this is only 1\% for MobileNet. 
\begin{table}[]
	\centering
	\begin{minipage}[b]{1.0\linewidth}
	\caption{Performance of distributed inference with model compression for MobileNet-V1 using 3 Raspberry Pi 4, $r=42$ M{\upshape bps}}
	\centering
	\resizebox{0.9\textwidth}{!}{
		\begin{tabular}{c|ccc}
			\toprule
		Model & $T$ (ms) & $T_{\text{HALP}}$ (ms) & Gain\\ 
			\cline{1-4}
  MobileNet\_v1\_1.0\_224        & 1739 &   1078 & 1.61\\ 
  MobileNet\_v1\_1.0\_192        & 1603 &   924 & 1.73\\ 
  MobileNet\_v1\_1.0\_160        & 1317 &   804 & 1.63\\ 
  MobileNet\_v1\_0.75\_224        & 1442 &   907 & 1.58\\ 
  MobileNet\_v1\_0.75\_192      & 1126 &   718 & 1.56\\ 
  MobileNet\_v1\_0.75\_160        & 1049 &   668 & 1.57\\ 
  MobileNet\_v1\_0.50\_224        & 1126 &   700 & 1.60\\ 
  MobileNet\_v1\_0.50\_192      & 959 &   593 & 1.61\\ 
  MobileNet\_v1\_0.50\_160       & 749 &   462 & 1.62\\ 
  MobileNet\_v1\_0.25\_224       & 689 &   432 & 1.59\\ 
  MobileNet\_v1\_0.25\_192        & 617 &  388 & 1.59\\ 
  MobileNet\_v1\_0.25\_160       & 555 &   350 & 1.58\\ 
\bottomrule
\end{tabular}}%
\label{Tab:GainMobileNet}
\end{minipage}
\end{table}%
Fig.~\ref{Fig:accuracyVStime} shows the trade-off between accuracy and inference time of the 12 lightweight MobileNet-V1 models on ImageNet dataset. Fig.~\ref{Fig:accuracyVStime} also includes the performance of VGG-16 and we can see that VGG-16 improves 1\% in accuracy but uses much more time for inference.
In general, distributed inference with task partitioning can achieve higher accuracy for a given deadline for inference. For example, under the inference time constraint 1000ms, distributed inference with task partitioning can achieve an accuracy of 71\%, whereas the conventional method without task partitioning can achieve an accuracy of 65\% by running a model with higher prunning rate.
\begin{figure}[t]
  \centering
  % include first image
  \includegraphics[width=0.9\linewidth]{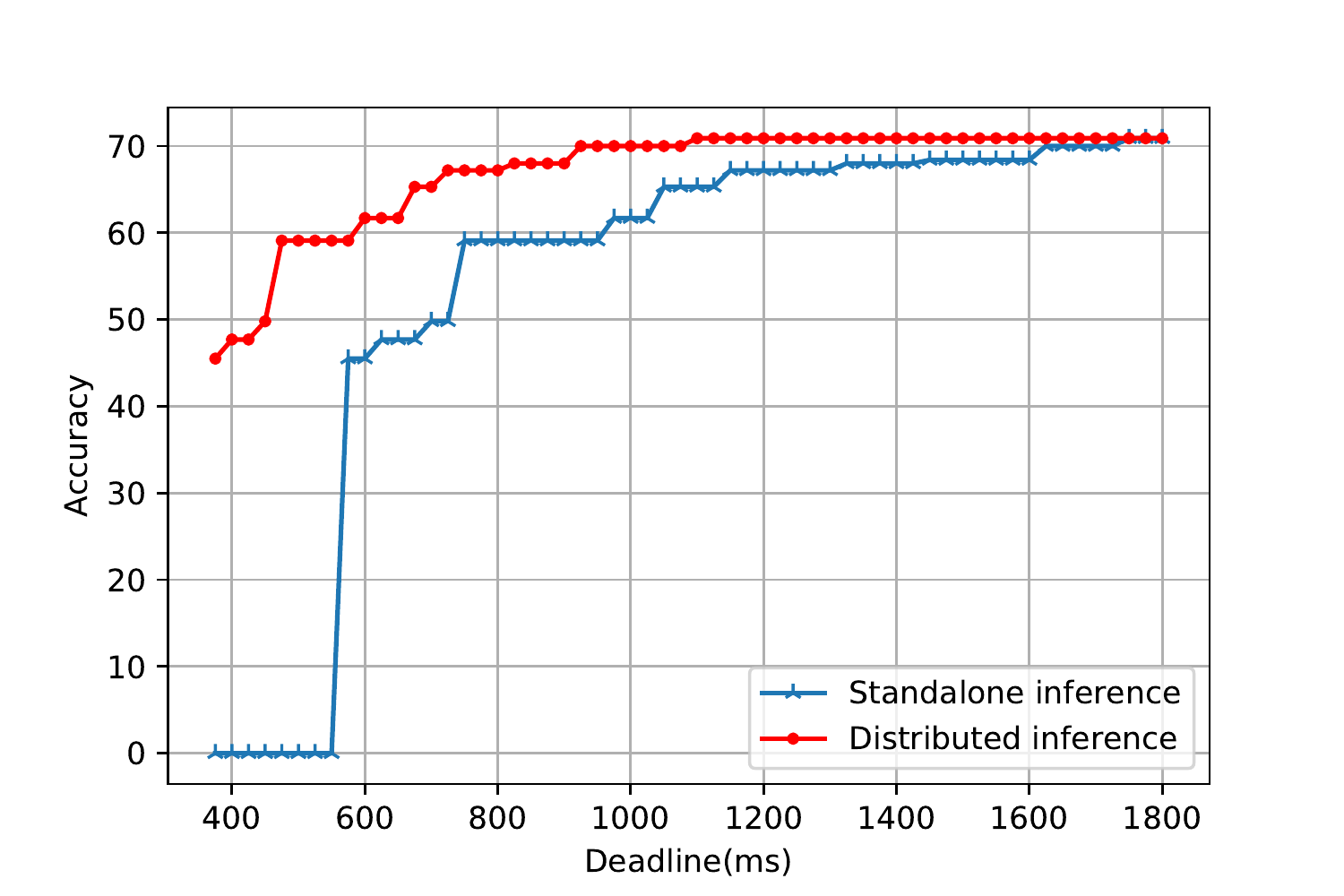}  
 \caption{Inference accuracy under different deadlines on raspberry Pi 4}\label{Fig:DeadlineVsAcc}
\end{figure}

\subsection{Dynamic model selection with distributed inference}~\label{Sec:ModelSelection}
Time-critical IoT applications using CNN are required to complete inference within latency constraints, otherwise, the inference result is often useless and the task is regarded as a failure. Service reliability in such applications depends on inference accuracy achieved within the latency constraints. As discussed in Section \ref{model compression}, the hyperparameters of the lightweight models affect the trade-off between latency and accuracy. Therefore, to maximize service reliability, we can dynamically select the optimum one among all the available models in use based on the input image size and the current data throughput to strike a balance between latency and accuracy. 

We conduct simulations to study how dynamic model selection with distributed inference HALP performs compared to the stand-alone computation (i.e. without task partitioning). 
%As each line in Fig.~\ref{Fig:accuracyVStime} is independent and separate, our idea is very straightforward, for each task with image size, the required deadline, and the current channel condition, we select the model with the highest accuracy from all the models that meet the deadline.
For the stand-alone computation, an image is taken and computed by the host ED itself, hence, the channel state and image size do not affect the final inference
time, as long as the inference time is less than the deadline, a model is qualified. For distributed inference, the total inference time includes the image offloading time and inference time; hence, the data throughput and image size need to be considered.

Fig.~\ref{Fig:DeadlineVsAcc} shows the highest accuracy achieved on ImageNet using model selection with distributed inference HALP and model section with stand-alone computation at different deadlines. We vary the latency constraint from 375 ms to 1800 ms. Since the shortest inference time using the stand-alone method is 555 ms (i.e. MobileNet-V1 with $\alpha =0.25 $ and $\rho = 160 \times 160$), when the latency constraint is less than 555 ms, the stand-alone method is not able to meet the deadline. This is the reason that the accuracy of the stand-alone method is always 0 when the deadline is below 555 ms. One way to complete an inference task within 555 ms is to use distributed inference HALP. It can be seen that the accuracy of HALP is higher than that of the stand-alone method for any given deadline, in particular, when the latency constraint is below 555 ms. The accuracy difference between distributed inference and stand-alone method gradually decreases as the latency constraint gets relaxed. Distributed inference reaches its maximum accuracy at 1100 ms when using VGG-16, while the stand-alone method needs 1750 ms.

\begin{figure}[t]
  \centering
  % include first image
  \includegraphics[width=0.8\linewidth]{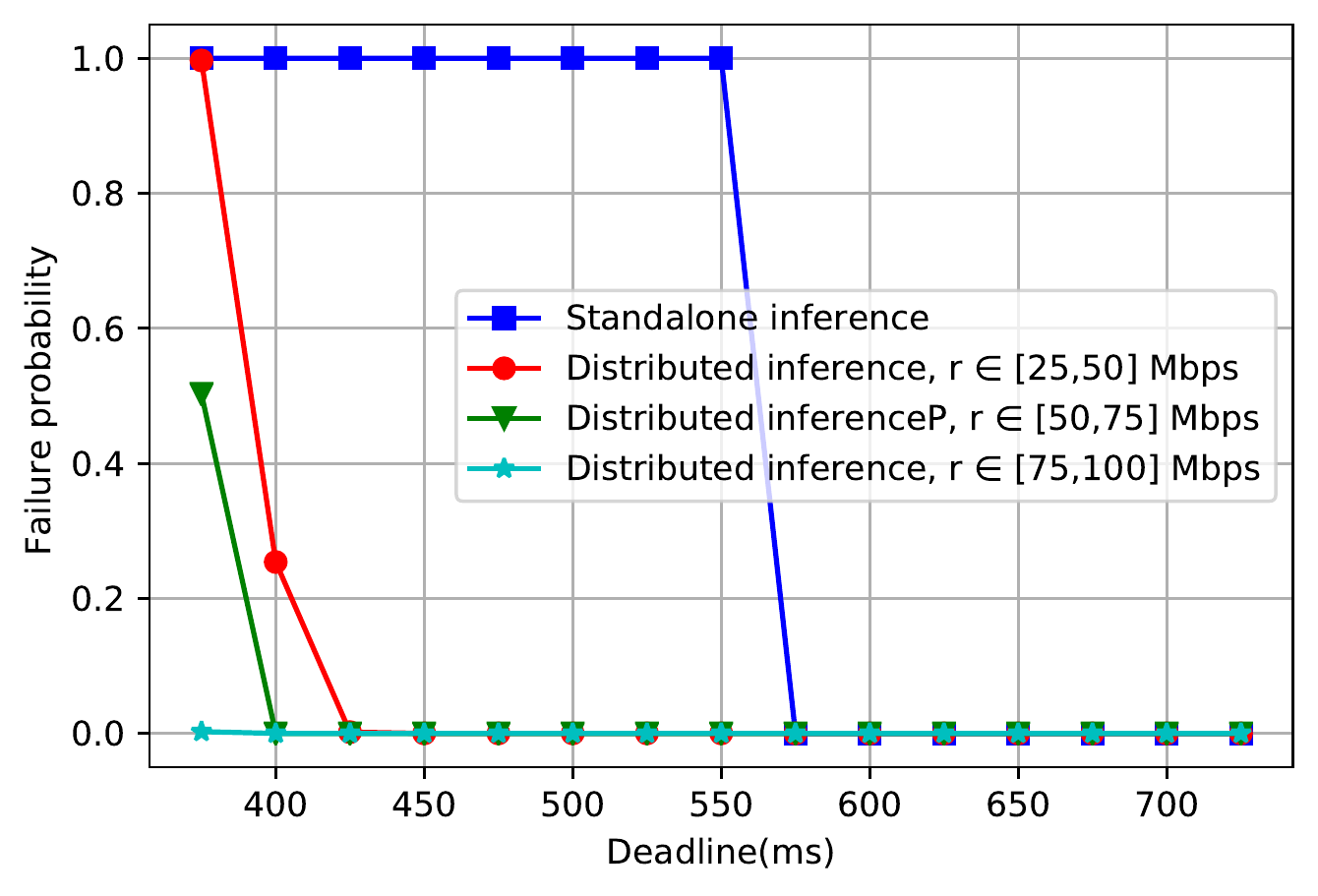}  
 \caption{Failure probability under different deadlines on raspberry Pi 4}\label{fig:FailPro}
\end{figure}

We further study the effect of transmission latency on the failure probability of the task. The image size of the arriving tasks follows a Gaussian distribution with a mean of 300 KB and a variance of 50 KB. We consider three different channel conditions, data throughput is evenly distributed between 25 Mbps to 50 Mbps (poor state), 50 Mbps to 75 Mbps (medium state), 75 Mbps to 100 Mbps (good state). For each deadline, we randomly generated 10,000 tasks. 

Fig.~\ref{fig:FailPro} shows failure probability under different deadlines. The failure probability is defined as the number of tasks that fail to meet the deadline divided by the total number of tasks. As explained previously, the stand-alone method cannot complete a task within a deadline smaller than 555 ms, hence the failure probability is 100\% for any deadline below 555 ms, but when the deadline is above 555 ms, it is able to handle all tasks. With a deadline of 375, when the channel state is poor, for distributed inference using HALP, the failure probability is approaching to 1 and it is reduced to about 50\% when the channel state is moderate, and almost all tasks can meet this deadline when the channel state is good. When the deadline is 425 ms, all tasks can meet the deadline for all channel states using distributed inference. 

\begin{figure}[t]
  \centering
  % include first image
  \includegraphics[width=0.8\linewidth]{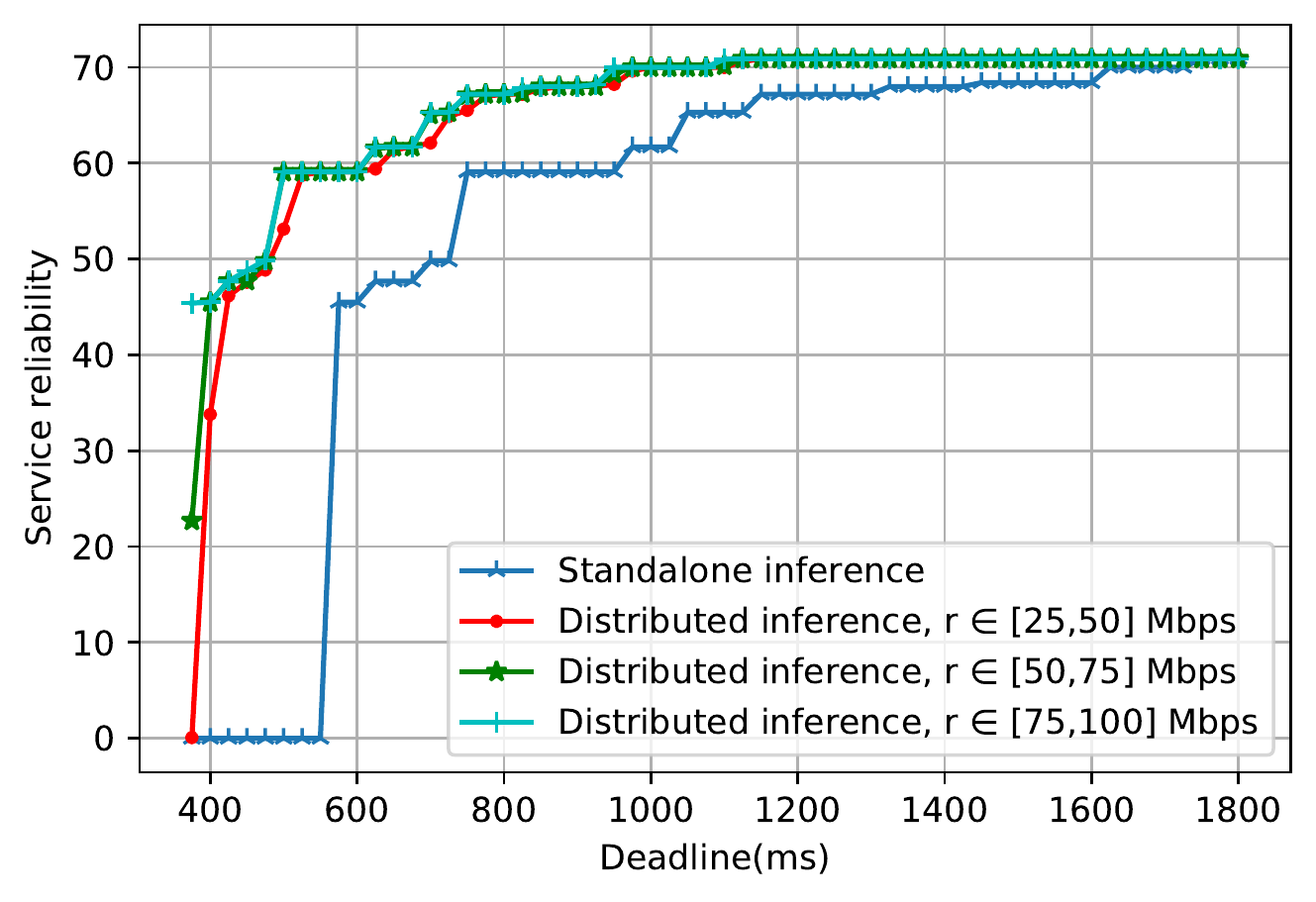}  
 \caption{Service reliability under different deadlines on raspberry Pi 4}\label{fig:ServiceReliability}
\end{figure}

To evaluate the service reliability which is defined as inference accuracy multiplied by the success probability (i.e. 1-failure probability).  
%\[ QoS = Average\:Accuracy\times (1-Failure\;Rate)\]
Fig.~\ref{fig:ServiceReliability} shows the service reliability under different deadlines. %There is no difference between the service reliability and the accuracy of the standalone method. 
In general, the service reliability of distributed inference HALP is higher than that of the stand-alone method, as distributed inference not only accelerates inference but also allows to use a more sophisticated CNN model to achieve higher inference accuracy within the deadline. This shows that dynamic model selection with distributed inference HALP is a promising solution for time-critical IoT applications. The improvement using distributed inference is significant when a task fails to meet a given deadline. When the deadline is relaxed, the improvement gets smaller. For distributed inference, it can also be seen that poor channel state has a negative impact on service reliability, but when the data rate is over a certain threshold, continuing to increase the data rate will not increase service reliability further, meaning it reaches the maximum service reliability, i.e. the inference accuracy of the CNN model. 

\section{Conclusion}~\label{Sec:Conclusion}
We implement a working prototype of a distributed inference method, HALP, for both VGG-16 and MobileNet-V1 using three Raspberry Pi 4. For VGG-16, it can achieve $1.71 \times$ in inference time. Then we combine the distributed inference with CNN model compression, in other words, we partition tasks and let the lightweight CNN models compute each sub-task. This can further reduce inference time but at the cost of accuracy loss. To strike a balance between latency and accuracy, we apply dynamic model selection. It shows that dynamic model selection with distributed inference HALP is effective in maximizing service reliability under latency constraints.

%We adjusted the task partitioning ratio based on the segment-based partitioning so that the original small and fixed overlapped area becomes more balanced. Experiment results show that the balanced partitioning method further accelerates the inference by 20\%. In addition to VGG16, we also applied HALP to the task of MobilenetV1, we achieved an average acceleration of 1.61x on MobileNetV1 task, which is slightly better than 1.5x on the VGG16 task. In order to deal with tasks with hard deadlines, we took 12 different versions of MobilenetV1 made from the cross product of width multiplier $\alpha \in \{1, 0.75, 0.5, 0.25\}$ and resolutions $\{224\times 224, 192\times 192, 160\times 160\}$, we partition the tasks of these pruned models, and then proceed to model selection. For each incoming task, by considering the desired accuracy and inference time, we calculate the optimal setup of the network, simulation results show that using HALP significantly improves service quality compared to the standalone method.
\section*{Acknowledgment}
This work is supported by Agile-IoT project (Grant No. 9131-00119B) granted by the Danish Council for Independent Research.

\bibliographystyle{IEEEtran}
\bibliography{conference_101719}
\end{document}